\documentclass{article}
\usepackage{ijcai19}

\usepackage{times}
\usepackage{soul}
\usepackage{url}
\usepackage[hidelinks]{hyperref}
\usepackage[utf8]{inputenc}

\usepackage[small]{caption}
\usepackage{graphicx}

\usepackage{amsmath}
\usepackage{booktabs}
\usepackage{times}
\usepackage{latexsym}
\usepackage{subcaption}
\usepackage{comment}
\usepackage{times}
\usepackage{soul}
\usepackage{booktabs}
\usepackage{url}
\usepackage{amsmath}
\usepackage{lipsum}

\usepackage{amssymb}
\usepackage{booktabs}
\usepackage{algorithm}
\usepackage{algorithmic}
\usepackage{multirow}
\usepackage{multicol}
\usepackage{subcaption}
\usepackage{caption}

\newcommand\blfootnote[1]{%
  \begingroup
  \renewcommand\thefootnote{}\footnote{#1}%
  \addtocounter{footnote}{-1}%
  \endgroup
}
\usepackage[symbol]{footmisc}

\renewcommand{\thefootnote}{\fnsymbol{footnote}}

\usepackage{fontawesome}

\makeatletter
\newcommand{\printfnsymbol}[1]{%
  \textsuperscript{\@fnsymbol{#1}}%
}
\makeatother
\title{Hindi Question Generation Using Dependency Structures \footnote{\llap{\textsuperscript{*}}This work was presented at 2nd Workshop on Humanizing AI (HAI) at IJCAI'19 in  Macao, China.}}

\author{Kaveri Anuranjana\printfnsymbol{2}\and
Vijjini Anvesh Rao\printfnsymbol{2} \and 
Radhika Mamidi\\
\affiliations
Language Technologies Research Centre (LTRC) \\
Kohli Center on Intelligent Systems (KCIS) \\
International Institute of Information Technology, Hyderabad, India \\
\{kaveri.anuranjana,vijjinianvesh.rao\}@research.iiit.ac.in\\radhika.mamidi@iiit.ac.in}

\begin{document}
\maketitle
\begin{abstract}

Hindi question answering systems suffer from a lack of data. To address the same, this paper presents an approach towards automatic question generation. We present a rule-based system for question generation in Hindi by formalizing question transformation methods based on karaka-dependency theory. We use a Hindi dependency parser to mark the karaka roles and use IndoWordNet a Hindi ontology to detect the semantic category of the karaka role heads to generate the interrogatives. We analyze how one sentence can have multiple generations from the same karaka role's rule. The generations are manually annotated by multiple annotators on a semantic and syntactic scale for evaluation. Further, we constrain our generation with the help of various semantic and syntactic filters so as to improve the generation quality. Using these methods, we are able to generate diverse questions, significantly more than number of sentences fed to the system.
\blfootnote{\printfnsymbol{2}These authors have contributed equally to this work}
\end{abstract}

\section{Introduction}
Neural networks have been able to make the learning jump from simple question answering tasks like \cite{VoorheesTrec-8} to complicated tasks like non-factoid question answering \cite{habernal2016,keikha2014} (while factoid questions focus on concise answers, non-factoid question answering can cover lengthier answers). While hand-crafted datasets are rich in information \cite{richardson2013mctest}, their quantity is often not enough for application of machine learning methods like deep neural networks \cite{trecsota}. Recent improvements in neural network architectures \cite{attentionoverattention,hu2017reinforced,wang2017gated} have beaten human performance in an increasing challenging task of non-factoid question answering \cite{squad1}. This has been in part possible due to the large size of the dataset available for such tasks \cite{squad2,yang2015wikiqa,Hermann2015Cloze}.

For Indian languages however, resource scarcity has been a big challenge. For machine learning based solutions known for handling complex sentence structures, it is imperative to prioritize data collection. Due to the plethora of Hindi resources in news and Wikipedia knowledge domains, immense potential lies in question generation for providing large scale question answering datasets. In this paper we provide a dependency parser and semantic word annotations based, rule-based question generation framework for various types of questions' automated generation. As our model is fully rule-based, it does not have to rely on any previously labelled data, hence overcoming the problem of resource scarcity. We evaluate our model by native Hindi speakers, who rate our generation at a semantic and syntactic level independently.

\section{Related Work}

\subsection{Question Generation}
Previous works on question generation relied on templates \cite{mostow2009generating,sneiders2002automated}. Further work was done in neural generation as well. Generating factoid questions with neural networks \cite{factoid_rnn_ques_generate,AQG_deeprl} with a sizeable corpus was done. However neural generation methods require sizeable amount of corpus to train machine learning models. Hence, it is difficult for such models to facilitate Hindi, which is rather resource scarce in this regard.

\subsection{Hindi Question Answering}
With limited work done for them by \cite{7514599,sekine2003hindi}, question answering in Indian languages has a very poor representation. Among these languages, Hindi has been most extensively researched, yet few rule-based QA systems have been proposed \cite{kaurautomatic,yogish2017survey}. \cite{khyati_webshodh}'s system works on code mixed Hindi-English data. \cite{weston2015babi}'s work on a Hindi QA corpus generation was originally done in English. The Hindi version was artificially created through translation methods. These methods do not involve human annotated Hindi data.

\subsection{Hindi Dependency Parsing}
Dependency parsing in Indian languages follows Indian grammatical tradition by Panini, an early $4^{th}$ century BC linguist. 
According to Paninian Grammar, Hindi dependency roles can be explained in terms of Karakas which can loosely correlate to typical dependency labels used in English. The head of a sentence is the main verb while the rest of the phrases are children of the main verb. The role of these child nodes with respect to the main verb is the Karaka rols. 
\begin{figure}
    \centering
    \includegraphics[width=8cm]{{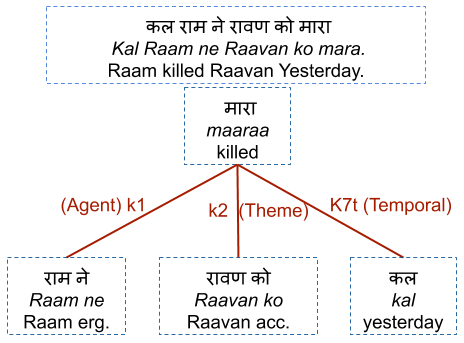}}
    \caption{Karaka dependency labels for the sentence ``Kal Raam ne Raavan ko mara.''}
    \label{fig:deptree}
\end{figure}
For example \textit{``Kal Raam ne Raavan ko mara.''}(as shown in Figure \ref{fig:deptree}) is translated as \textit{``Raam hits Raavan yesterday.''} Here, the relations are specified as - \textit{Raam} which is \textit{k1 karaka} or doer, \textit{Raavan} which is \textit{k2 karaka} or patient of the verb \textit{``hits''} and \textit{kal} which is \textit{k7t karaka} the time this action took place. Karakas sometimes are expressed with their case markers (\textit{``ne''} and \textit{``ko''} in this case).
\cite{begum2008dependency} describes the annotation scheme of these dependency parsers. In the following sections, we explain rules we developed for each karaka role along with how the karaka could be understood in terms of thematic roles.\footnote{We use the dependency parser for Hindi provided at https://bitbucket.org/iscnlp/parser}
\section{Experiment}
In this section we talk about the rules used for our generation and experimental setup. Dependency labels are based on \cite{begum2008dependency}. In many of our rules, a single rule may dictate more than one question generation. This lets us maximize our number of generations. For evaluating our rule-based system's performance, we take sentences from online learning educational websites \footnote{\protect\url{https://sandeepbarouli.com/}}\footnote{\protect\url{http://www.2classnotes.com/}}. In total we were able to generate 112 questions from just 30 selected sentences. More details regarding our results are explained in Section \ref{sec:resultsanddiscussion}. 

Question generation in our system often relies on the case of the keyword. Case in Hindi, Oblique or Direct, is determined by whether the role of keyword was assigned by an explicit case marker, for example the postpositions \textit{ke, ko, se} or \textit{ka} (Oblique Case) or directly by the verb (Direct Case).

\subsection{k1 Case}
Definition: Doer or Patient\\
Generally the subject of the verb. We substitute with direct or oblique interrogative words based on whether the word assigned k1 is direct or oblique (direct case does not have a postposition between the noun whereas oblique case does).
\begin{equation}
\begin{aligned}
    & \textit{\text{\underline{X (ne)} Y ko khaya/kha raha hai.}} \longrightarrow \\
    & \text{\underline{X} is eating Y.} \\
    & \begin{cases}
        \textit{\text{\underline{kaun} Y ko kha raha hai ?}}\\
        \textit{\text{\underline{kisne} Y ko khaya ?}} \\
        \text{\underline{who} is eating Y ?}
    \end{cases}
\end{aligned}
\end{equation}

\subsection{k1s Case}
Definition: Copula\\
If X is a property of Y (\textit{eg- ``raam geela hai.''} translated as \textit{``Ram is wet.''} has the corresponding question, \textit{``ram kaisa hai ?''} translated as \textit{``In what condition is Ram ?''}), then the interrogative is \textit{``kaisa''} or \textit{``in what condition''}. Whereas, if X is a occupation role for Y (\textit{eg- ``raam ek dauctar hai.''} translated as \textit{``Ram is a doctor.''} has the corresponding question \textit{``raam kaun hai ?''} or \textit{``Who is Ram?''}), then the interrogative is \textit{``kaun''} or \textit{``who''}.

In our implementation, we use IndoWordnet \cite{bhattacharyya2017indowordnet}, a WordNet based ontology in Hindi to define those words which fall in the Human category as a ``occupation'' and the other words as ``Property''.
\begin{equation}
\begin{aligned}
    \textit{\text{X \underline{Y} hai.}}
    &\longrightarrow \\
    \text{X is \underline{Y}.} \\
    &\textit{\text{X \underline{kaun/kaisa} hai ?}}\\
    &\text{\underline{What/How} is X ?}
\end{aligned}
\end{equation}

\subsection{k2 Case}
Definition: Recipient, Patient or Beneficiary \\
Just like k1, we substitute with direct or oblique interrogative words based on whether the keyword is direct or oblique.
This karaka is overloaded (multiple outputs are possible that are not in free variation) due to direct and oblique case (defined by presence or absence of case marker). 
\begin{equation}
\begin{aligned}
    & \textit{\text{X \underline{Y (ko)} khaataa hai.}} 
    \longrightarrow \\
    & \text{X eats Y.} \\
    & \begin{cases}
        \textit{\text{X \underline{kisko} khaataa hai ?}} \\ 
         \text{X eats \underline{whom} ?} \\
        \textit{(with postposition; generally a living object)}\\
        \textit{\text{X \underline{kya} khaataa hai ?}} \\ 
        \text{X eats \underline{what} ?} \\ 
        \textit{(without postposition; generally non-living)}
    \end{cases}
\end{aligned}
\end{equation}

\subsection{k2p Case}
Definition: Goal, but related to location \\
We delete the keyword along with its postposition and substitute it with a locative interrogative. As multiple such interrogatives are possible, we generate multiple questions from a single input in this rule. Both the generations are in free variation.
\begin{equation}
\begin{aligned}
    \textit{\text{X \underline{Y (ko)} gayi thi.}}  
    & \longrightarrow\\ 
    \text{X went to \underline{Y}.}\\
   &\begin{cases}
        \textit{\text{X \underline{kidhar} gayi ?}}\\
        \textit{\text{X \underline{kahan} gayi ?} }\\
        \text{\underline{Where} did X go?}
    \end{cases}
\end{aligned}
\end{equation}

\subsection{k3 Case}
Definition: Instrument or Path\\
Two case markers - \textit{se} and \textit{ke dwaaraa} are possible in indicating this role. The interrogatives become \textit{kisse} and \textit{kiske dwaaraa} respectively. If k3 specifies the Path (detected via IndoWordNet labels) through which the action happened, \textit{kisse} and \textit{kisse hokar} are used as interrogatives.
\begin{equation}
\begin{aligned}
    &\textit{\text{X \underline{Y se/ke dwaaraa} jaati hai.} }
    \longrightarrow \\
    & \text{X goes \underline{with/via Y}.} \\
    & \begin{cases}
       \textit{ \text{X \underline{kisse/kiske dwaaraa} jaati hai ?}}\\
        \textit{\text{X \underline{kisse/ kisse hokar} jaati hai ?}}\\
        \text{X goes \underline{through/with what} ?}
    \end{cases}
\end{aligned}
\end{equation}

\subsection{rt Case}
Definition: Purpose \\
Postposition \textit{ke liye} (for) is used to indicate this role. If Y is a human then replacement is done with \textit{kiske liye} (for whom), otherwise with \textit{kyon}(why). Y's category as person or not is decided from IndoWordnet.
\begin{equation}
\begin{aligned}
     \textit{\text{X ne \underline{Y ke liye} likha.}} 
     & \longrightarrow \\
    \text{X wrote for Y.} \\
    &\begin{cases}
       \textit{ \text{X ne \underline{kiske liye} likha ?}}\\
         \text{X wrote \underline{for whom} ?}\\
     \textit{   \text{X ne \underline{kyon} likha ?}}\\
         \text{X wrote \underline{why} ?}\\
    \end{cases}
\end{aligned}
\end{equation}
\subsubsection{rh Case}
Definition: Reason \\
Whenever parser detects because term \textit{kyunki} (because), it largely gives rh label to the consequent chunk, the reason. We replace the entire chunk along with because term with the corresponding question word. 
\begin{equation}
\begin{aligned}
    \textit{\text{X ne Z \textit{kiya} \underline{kyunki Y}.}} 
   &\longrightarrow\\
    \text{X did Z \underline{because Y}.} \\
 &   \textit{\text{X ne Z \underline{kyon} \textit{kiya} ?}} \\
&    \text{X did Z \underline{why} ?} \\
\end{aligned}
\end{equation}

\subsection{k5 Case}
Definition: Source \\
If Y is a place, \textit{kahan/kidhar} (from where) is more the appropriate replacement. Semantically, both interrogatives carry different meanings. \textit{kisse} implies something was separated from Y where Y could be human or thing, whereas \textit{kahan/kidhar} carries locative information. \textit{kahan/kidhar} are in free variation. 
\begin{equation}
\begin{aligned}
    \textit{\text{X \underline{Y se} bhaagaa.}}
   & \longrightarrow \\
     \text{X ran \underline{from Y}.} \\
    & \begin{cases}
        \textit{\text{X \underline{kisse} bhaagaa} ?}\\
        \text{X ran \underline{from whom/what} ?} \\
        \textit{\text{X \underline{kahan/kidhar se} bhaagaa ?}} \\
        \text{X ran \underline{from where} ?}\\
    \end{cases}
\end{aligned}
\end{equation}

\subsection{r6 Case}
Definition: Possession \\ 
k6 isn't a karaka in the annotation schema proposed by \cite{begum2008dependency} however, the parser marks r6 as the possession relation (genitive case). The interrogative root form, \textit{kis-} is modified with the suffix derived from the postposition (\textit{-ka / -ke / -ki}).
\begin{equation}
\begin{aligned}
    & \textit{ \text{\underline{X (-ka / -ke / -ki)} Y ja raha/rahe/rahi hain.}}
     \longrightarrow \\
    & \text{\underline{X's} Y is going.} \\
    & \textit{\text{\underline{kiska/kiske/kiski} Y ja rahe hain ?}} \\
    & \text{\underline{Whose} Y is going ?}
\end{aligned}
\end{equation}
Moreover, if the object Y is a non-living object then we can construct another question from r6 with modification to the verb according to the object's gender (as shown in italics below).
\begin{equation}
\begin{aligned}
    &\textit{\text{X \underline{ka Y} \textit{ja raha hain}.}}
     \longrightarrow \\
    &\text{\underline{X's} Y is going.} \\
    &\textit{\text{X \underline{ki kaun si vastu} \textit{ja rahi hain} ?}} \\
    &\text{\underline{X's what item} is going ?}
\end{aligned}
\end{equation}

\subsection{k7s Case}
Definition: Spatial Locative \\
Multiple interrogative words are possible in free variation in this case. Just like other rules. we consider all possibilities to supplement the data size. 
The substitution rule is as follows for two postpositions :
\begin{equation}
\begin{aligned}
    & \textit{\text{X \underline{Y mein/par} baithi thi.}}
        \longrightarrow \\
    & \text{X is sitting \underline{in/on Y}.} \\
    & \begin{cases}
    \textit{\text{X \underline{kahan/kidhar} baithi thi ?}}\\
    \text{X is sitting \underline{where} ?}\\
    \textit{\text{X \underline{kis mein/kis par} baithi thi ?}} \\ 
    \text{X is sitting \underline{in/on} what ?}
    \end{cases}
\end{aligned}
\end{equation}

\subsection{k7t Case}
Definition: Temporal Locative \\
Multiple interrogative phrases are possible if the keyword is a date. However lack of differentiation between day as compared to time we consider all generations. This leads to larger generation set, but at the cost of quality of generations. We talk more about overgeneration in Section \ref{sec:overgeneration}. The substitution rule is as follows:
\begin{equation}
\begin{aligned}
    \textit{\text{X \underline{Y ko} jaegi.} }
    & \longrightarrow\\
    \text{X will go \underline{on Y}.} \\
    &\begin{cases}
       \textit{ \text{X \underline{kab} jaegi ?}}\\
        \text{X will go \underline{when} ?}\\
        \textit{ \text{X \underline{kis/konse din} jaegi ?}}\\
        \text{X will go \underline{which day} ?}
    \end{cases}
\end{aligned}
\end{equation}

\section{Results and Discussion}
\label{sec:resultsanddiscussion}
This section describes the results obtained with the carried experimen. First, we discuss how the questions generated by the system are graded by human annotators. Further, we discuss the results obtained from the generation approach.

\subsection{Results}
Five Hindi native speakers were told to rate the generated questions on a five point Likert Scale for grammatical correctness (Syntax) and meaningfulness (Semantic). Alongside the questions, they were also shown the original sentences for reference. The mean and median for each of the karaka-rules and overall score are reflected in the table \ref{results}.

For a total of 30 sentences of varying difficulty selected from the Hindi corpus, we ran the generation rules for each of the karakas. The generator returned more than 100 questions from a much smaller set, which indicates a case of over-generation. The k1-rule (a frequently occurring linguistic phenomena) on itself generates 30 questions alone. This might be because every sentence generally has a k1 Karaka i.e. Doer or Agent. Similar karaka-wise distribution of the number of questions generated, along with the human scores are shown.
This over-generation can be exploited for data augmentation where the training set size can be expanded in magnitudes. As the generator generates from questions from multiple semantic roles, the augmented dataset can be much richer in terms of the semantic diversity of the questions. 

\subsection{Discussion}
\label{sec:discussion}
The annotators were also told to point out, if they observed blatant non-grammatical errors. We discuss some of them below.\\

\noindent\textbf{Q1:} \textit{hawaa ne sooraj se \underline{kaha} - kaun tumse adhik balwaan hai ?}\\
\textbf{T1:} The wind \underline{told} the sun - who is mightier than you ?\\
Semantic transfer of words like ``told'' to ``asked'' need to be made when generating questions. \\

\noindent\textbf{Q2:} \textit{hawaa ne kaha - \underline{kya} ?} \\
\textbf{T2:} The wind said \underline{what}\\
While writing the rules, we had considered the most frequently occurring Subject-Object-Verb (SOV) word-order as the case above suggests. Transformation rules to bring the interrogatives at the right position need to be applied in such cases. The correct form in this example should have been \textit{``hawaa ne \underline{kya} kaha ?''}\\

\noindent\textbf{Q3a:}\textit{ hawaa ne kya \underline{dikhani} shuru \underline{ki} ?}\\
\textbf{T3a:} \underline{What(\textit{msc.})} did the wind start \underline{showing(\textit{fem.})} ? \\
\textbf{A3b:} \textit{hawaa ne apni taakat \underline{dikhani} shuru \underline{ki} }.\\
\textbf{T3b:} The wind started to \underline{show(\textit{fem.})} it's power(\textit{fem.}).\\
The gender of verbs in Hindi agree with the direct object i.e. the word power(msc.). However the interrogative \textit{``kya''} is masculine, hence the simple substitution method of the generator fails. The verb \textit{``ki''} needs to be modified accordingly to \textit{``kiya''} to make the question - \textit{``hawaa ne kya \underline{dikhana} shuru \underline{kiya}'' ?}.

\begin{table}
\centering
\begin{tabular}{|l|l|l|}
\hline
& Before Pruning & After Pruning \\ 
     \hline
Semantic  & 3.019&3.244 \\ 
Syntax & 3.336&3.726 \\
\hline
 Count  & 112&68  \\ 
\hline

\end{tabular}
\caption{Mean scores and number of generations before and after avoiding overgeneration (pruning)}
\label{tab:after}
\end{table}

\section{Overgeneration} 
\label{sec:overgeneration}
As we generously considered every karaka marked by the parser for generation, bad quality generations were also generated. They could have been due to the fault in the rules, or parser errors. During the annotation procedure, the native speakers were asked to point out blatant incorrectness of the questions generated and collected them. We classify them into surface-level (mostly morphological features), syntactic and semantic faults and provide an analysis.

\begin{table}
\centering
\begin{tabular}{|l|l|l|l|l|}
\hline
Karaka & Criteria & Mean & Median & Count  \\ 
\hline
\multirow{2}{*}{k1}      & Syntax         & 4.28 & 5      & \multirow{2}{*}{30}      \\
     & Semantic         & 3.76 & 4      &        \\ 
\hline
\multirow{2}{*}{k1s}       & Syntax         & 2.24 & 2      & \multirow{2}{*}{7}      \\
     & Semantic         & 2.24 & 2      &        \\
\hline
\multirow{2}{*}{k2}      & Syntax         & 2.84 & 3      & \multirow{2}{*}{17}      \\
     & Semantic         & 3.52 & 3      &        \\ 
\hline
\multirow{2}{*}{k2p}       & Syntax         & 3.2 & 3      & \multirow{2}{*}{2}      \\
     & Semantic         & 4.2 & 4      &        \\
\hline
\multirow{2}{*}{rt}      & Syntax         & 2.05 & 2      & \multirow{2}{*}{4}      \\
     & Semantic         & 2 & 2      &        \\ 
\hline
\multirow{2}{*}{rh}       & Syntax         & 2.4& 2      & \multirow{2}{*}{1}      \\
     & Semantic         & 3& 2     &        \\
\hline
\multirow{2}{*}{k5}      & Syntax         &3.2 & 4      & \multirow{2}{*}{6}      \\
     & Semantic         &    3.72    & 5    &   \\ 
\hline
\multirow{2}{*}{r6}       & Syntax         &3.48 &    4   & \multirow{2}{*}{13}      \\
     & Semantic        &    4.08    &    4  &  \\
\hline
\multirow{2}{*}{k7t}      & Syntax         &3.5 &    3 & \multirow{2}{*}{14}      \\
     & Semantic          &   3.2   & 4 &      \\ 
\hline
\multirow{2}{*}{k7p}       & Syntax         & 3&    3  & \multirow{2}{*}{18}      \\
     & Semantic         & 3.64& 5     &        \\
\hline
\multirow{2}{*}{\textbf{Total}}& Syntax& \textbf{3.019} &\textbf{3}& \multirow{2}{*}{\textbf{112}}      \\
     & Semantic  &\textbf{3.336} &    \textbf{4}  &        \\
\hline
\end{tabular}
\caption{The means and the medians of the 5-pt Likert scale for grammatical correctness (Syntax) and soundness in meaning (Semantic) of questions generated by each karaka rule, along with the number of generations from 30 input sentences.}
\label{results}
\end{table}

\subsection{Surface Level Filters}

\subsubsection{Anaphoras}
Questions lack anaphoras in them because unlike sentences from a paragraph, they lack context of surrounding sentences. Hence, values for anaphoras won't be cleared in questions, which are generally isolated. For this reason, removing questions which contain pronouns other than the interrogatives can lead to an increment in quality of generations.\footnote{We use the Parts-of-Speech tagger provided at https://bitbucket.org/iscnlp/pos-tagger}

\subsubsection{Gender Agreement}
\label{overgeneration_genderagreement}
In Hindi, intransitive verbs agree in gender, number and person with the subject. Verbs in questions however automatically get masculine gender. So when we construct questions, we would like to hide the gender of the answer. Hence, the verbs need to be converted to the general case, i.e., masculine.

For transitive verbs, if the subject is direct (does not have a case marker), the verb's morphological features like gender, number, etc. agree with the subject. But if the subject is oblique (has a case marker), the verb agrees with the object. If the subject is direct and the interrogative contains a case marker suffix in it and vice-versa, the verb has to be modified accordingly for agreement. Q3 from Section \ref{sec:discussion} illustrates an example for this.

We have dropped questions violating the agreement rules for now. Further modification of verbs and suffixes needs to be done to account for agreement.

\subsection{Syntatic Filters}

\subsubsection{Word Order}
Hindi is largely Subject-Object-Verb (SOV) order language and for cases as demonstrated in example Q2 in Section \ref{sec:discussion}, the order needs to be modified after substitution. For our experiment, we drop questions which do not follow the order.

\subsubsection{Questions}
Questions were selected as input sentences to the system. Words being tagged as karaka roles were substituted resulting in illegal constructions with multiple interrogatives. Such cases were removed. It stands to reason that the rules should not be applied to sentences that are already questions.

\subsubsection{Complex and Compound Sentences}
\label{overgeneration_complex}
Questions generally focus on seeking information about a particular topic, but non-simple sentences have phrases or clauses containing way more extra information. Question generation from sentences like these look very unnatural. Hence, we drop questions formed by such sentences by detecting the \textit{coof} tag which denotes conjunct of relation and take the length of the phrases on either side of the tag. If either side exceeds a threshold of $\theta$\footnote{For our experiment we took $\theta$ = 5} then we drop the sentence.

\subsection{Semantic Filters}
\label{overgeneration_semantic}
Semantic features need extensive resources and tools which may not be at disposal. Hence, we have not been able to implement such filters. One such case is Q1 from Section \ref{sec:discussion}.

\subsection{Improving question quality with filters}
The system overgenerates questions and as we have a generous quantity of generations, we apply the aforementioned filters to improve the quality of the generated dataset. We again calculate the performance with pruning overgenerated questions, shown in Table \ref{tab:after}. Overall we increased the quality by 0.2 to 0.4 point in Semantic and Syntax respectively. 

The annotators had also pointed out certain rules which result in the questions being modified and not dropped. These include the gender agreement rule in which we can use the IndoWordNet to modify the verbs accordingly which needs to be implemented in future work.

\section{Limitations and Future Work}
As the generator is driven by replacement of karaka roles with interrogatives, the changes are only surface level and a lot semantic changes to the question are lost. As mentioned in Section \ref{overgeneration_semantic}, semantic filters cannot be implemented with limited tools in Hindi and the filters discussed in Sections \ref{overgeneration_genderagreement} and \ref{overgeneration_complex} can be made into rules to generate better questions. As a design choice, we have considered only active voice rules, in future, we would like to include rules for passive voice as well. Doing so will increase the coverage of the sentences. 

The generator depends on the dependency parser to detect karakas from the parse trees. Hence any errors in the parser output (especially for longer sentences) propagates to the generator.

Hindi also acts as a free word order language based on presence or absence of case marker. Hence, transformation by moving the parse tree branches can generate more question-sentence pairs. Using active-passive voice transformation rules can further augment the dataset.

\section{Conclusion}
While rule-based systems are often considered out of time as compared to machine learning based methods, they don't rely on learning on huge corpora of data unlike the latter. This makes them effective for resource scarce languages. In this paper we present a question generation rule set for automated generation of Hindi questions from corresponding sentences. The question generation rules rely on information from pre-existing dependency parsers and word level ontologies for the Hindi language. Dependency parser outputs roles of every word in the sentence. These roles correlate with iterrogatives used in their corresponding questions.
Annotators manually evaluated the quality of our generations at a syntactic and semantic level. Furthermore we see that, a decrement in number of generations can lead to better quality generations. Hence, we give guidelines for avoiding overgeneration. 

\bibliography{ijcai19}
\bibliographystyle{named}

\end{document}